\def\year{2021}\relax
\documentclass[letterpaper]{article} 
\usepackage{aaai21}  
\usepackage{times}  
\usepackage{helvet} 
\usepackage{courier}  
\usepackage[hyphens]{url}  
\usepackage{graphicx} 
\usepackage{natbib}  
\usepackage{booktabs}
\usepackage{amsmath}
\usepackage{amssymb}
\usepackage{epsfig}
\usepackage{graphicx}
\usepackage{subcaption}
\usepackage{caption} 
\usepackage{color}

\title{An Attention Module for Convolutional Neural Networks}

\author{
    Zhu Baozhou\textsuperscript{\rm 1},
    Peter Hofstee\textsuperscript{\rm 1}\textsuperscript{\rm 2},
    Jinho Lee\textsuperscript{\rm 3},
     Zaid Al-Ars\textsuperscript{\rm 1}
    \\
}
\affiliations{

    \textsuperscript{\rm 1}Delft University of Technology, Delft, Netherlands \\
    \textsuperscript{\rm 2}
    IBM Research Austin, TX, USA\\
    \textsuperscript{\rm 3}
    Yonsei University, Seoul, Korea
}

\begin{document}

\maketitle

\begin{abstract}
  Attention mechanism has been regarded as an advanced technique to capture long-range feature interactions and to boost the representation capability for convolutional neural networks. However, we found two ignored problems in current attentional activations-based models: the approximation problem and the insufficient capacity problem of the attention maps. To solve the two problems together, we initially propose an attention module for convolutional neural networks by developing an AW-convolution, where the shape of attention maps matches that of the weights rather than the activations. Our proposed attention module is a complementary method to previous attention-based schemes, such as those that apply the attention mechanism to explore the relationship between channel-wise and spatial features. Experiments on several datasets for image classification and object detection tasks show the effectiveness of our proposed attention module. In particular, our proposed attention module achieves $1.00\%$ Top-1 accuracy improvement on ImageNet classification over a ResNet101 baseline and $0.63$ COCO-style Average Precision improvement on the COCO object detection on top of a Faster R-CNN baseline with the backbone of ResNet101-FPN. When integrating with the previous attentional activations-based models, our proposed attention module can further increase their Top-1 accuracy on ImageNet classification by up to $0.57\%$ and COCO-style Average Precision on the COCO object detection by up to $0.45$. Code and pre-trained models will be publicly available.
\end{abstract}

\section{Introduction}
Convolutional neural networks have demonstrated to be the gold-standard to solving various problems in the field of computer vision, including image classification \cite{he2016deep}, object detection \cite{liu2016ssd,ren2015faster}, and segmentation \cite{chen2017deeplab}. To improve their performance, many researchers explored various aspects of CNN design and implementation~\cite{gu2018recent}.

To enrich the representation power of CNNs, we can build substantially deeper convolutional neural networks. For example, VGGNet \cite{DBLP:journals/corr/SimonyanZ14a} stacks very small $3 \times 3$ convolutional layers, while ResNet~\cite{he2016deep} stacks residual blocks with skip connections. GoogLeNet~\cite{szegedy2015going} uses multi-scales of processing to capture spatial correlation. Wide ResNet \cite{BMVC2016_87} shows that the width increase of residual networks can enlarge the representation capability and reuse the features better. Xception \cite{chollet2017xception} and ResNeXt \cite{xie2017aggregated} expose new dimensions to increase cardinalities. Besides, recent literature \cite{jaderberg2015spatial, gregor2015draw, xu2015show} have investigated the attention mechanism since it can improve not only the representation power but also the representation of interests. Convolutional neural networks can extract informative features by blending cross-channel and spatial information \cite{hu2018gather}. Attention modules \cite{woo2018cbam, linsley2018learning} can learn "where" and "what" to attend in channel and space axes, respectively, by focusing on important features and suppressing unnecessary ones of the activations. Dynamic Filter Networks \cite{jia2016dynamic,li2019selective} generate the filters conditioned on the input and show the flexibility power of such filters because of their adaptive nature, which has become popular in prediction \cite{klein2015dynamic} and Natural Language Processing \cite{wu2018pay}. Both Dynamic Filter Networks and attention-based models are adaptive based on the inputs, but there are significant differences between them. Attention-based models \cite{hu2018gather,woo2018cbam} produce attention maps using the attention mechanism to operate on the activations of convolution. On the contrary, Dynamic Filer Networks \cite{su2019pixel,liu2019learning} generate input information-specific kernels, such as  position-specific kernels \cite{su2019pixel}, semantic label map-specific kernels \cite{liu2019learning}, and few-shot learning setting-specific kernels \cite{zhao2018dynamic}, which work as the weights of convolution. Our proposed attention module leverages the attention mechanism to compute the attention maps for attending the activations of convolution, so it is clear to categorized the models applied with our proposed attention module as attention-based models instead of Dynamic Filter Networks. 

Based on our analysis, the approximation problem and the insufficient capacity problem of the attention maps are ignored in current attentional activations-based models. Motivated by solving the two problems together, we develop an attention module and inspect the complementary relationship between our proposed attention module and previously published attention-based models, such as the attention augmented models and the attentional activations-based models. Our contributions are summarized as follows.
 
\begin{itemize}
    \item We point out and analyze two ignored problems of the current attentional activations-based models: the approximation problem and the insufficient capacity problem of the attention maps. To address the two problems together, we originally propose an attention module by developing an AW-convolution, where the shape of the attention maps matches that of the weights instead of the activations.
    \item Our proposed attention module is a complementary method to previous attention mechanism-based modules, such as Attention Augmented (AA) convolution \cite{Bello_2019_ICCV}, the SE \cite{hu2018squeeze} and CBAM \cite{woo2018cbam} modules in the attentional activations-based models. Integrating with our proposed attention module, the accuracy of AA-Net, SE-Net, and CBAM-Net will be improved further. 
    \item We use image classification and object detection tasks to demonstrate the effectiveness of our proposed attention module. With negligible computational complexity increase, our proposed attention module can boost the image classification and object detection task performance, and it can achieve better accuracy when integrating with other attention-based models.
\end{itemize}

\section{Related work}
In this section, we discuss the recent developments of network engineering and attention mechanism.
\subsection{Network engineering}
"Network engineering" has been one of the most active research areas since it targets building powerful convolutional neural networks on image classification, which are the backbones of various computer vision tasks and ensure their remarkable performance \cite{kornblith2019better}. Increasing the depth of convolutional neural networks has been regarded as an intuitive way to boost performance, which is the philosophy of VGGNet \cite{DBLP:journals/corr/SimonyanZ14a} and ResNet \cite{he2016deep}. In addition, since the skip connection from ResNet shows a strong ability to assist the gradient flow, WideResNet \cite{BMVC2016_87}, PyramidNet \cite{han2017deep}, Inception-ResNet \cite{szegedy2017inception}, and ResNeXt \cite{he2016identity,xie2017aggregated} are ResNet-based versions proposed to explore further the influence of the width, the increase of the width, the multi-scale and the cardinality of convolution, respectively. In terms of efficiency, DenseNet \cite{huang2017densely} reuses the feature maps by concatenating the feature maps from different layers. In particular, MobileNet \cite{DBLP:journals/corr/HowardZCKWWAA17,Howard_2019_ICCV} and ShuffleNet \cite{ma2018shufflenet} series present the advantage of depthwise convolution and the shuffle operation between various group convolutions, respectively. Another design approach uses automated neural architecture search, which achieves state-of-the-art performance regarding both accuracy and efficiency across a range of computer vision tasks \cite{tan2019mnasnet}.

\subsection{Attention mechanism}
The attention mechanism plays an important role in the human vision perceptron since it can allocate the available resources to selectively focus on processing the salient part instead of the whole scene \cite{rensink2000dynamic, corbetta2002control}. Multiple attention mechanisms are used to address a known weakness in convolution \cite{chen20182,szegedy2015going, ke2017multigrid, Chen_2019_ICCV, hu2018squeeze, linsley2018learning}, by capturing long-range information interactions \cite{DBLP:journals/corr/abs-1810-02019,NIPS2015_5866}. The Inception family of architectures \cite{szegedy2015going, szegedy2017inception}, Multigrid Neural Architectures \cite{ke2017multigrid}, and Octave Convolution \cite{Chen_2019_ICCV} aggregate the scale-space information, while Squeeze-and-Excitation Networks \cite{hu2018squeeze} and Gather-Excite \cite{hu2018gather} adaptively recalibrate channel-wise response by modeling interdependency between channels. GALA \cite{linsley2018learning}, CBAM \cite{woo2018cbam}, and BAM \cite{park2018bam} refine the feature maps separately in the channel and spatial dimensions. Attention Modules \cite{wang2017residual} and self-attention \cite{vaswani2017attention, Bello_2019_ICCV} can be used to exploit global context information. Precisely, non-local networks \cite{wang2018non} deploy self-attention as a generalized global operator to capture the relationship between all pairwise convolutional feature maps interactions. Except for applying the attention mechanism to computer vision tasks \cite{li2019zoom}, it has been a widespread adoption to modeling sequences in Natural Language Processing \cite{yang2019convolutional}.

\section{Proposed attention module}
In this section, we analyze the two ignored problems in current attentional activations-based models: the approximation problem and the insufficient capacity problem of the attention maps. To address the two problems together, we develop an attention module that mainly refers to the AW-convolution, where the shape of attention maps matches that of the weights rather than the activations. Besides, we refine the branch of calculating the attention maps to achieve a better trade-off between efficiency and accuracy. Last but not least, we integrate our proposed attention module with other attention-based models to enlarge their representational capability.

\subsection{Motivation}
First, we define basic notations in a traditional convolutional layer. In a traditional convolutional layer, the input activations, weights, and output activations are denoted as $I$, $K$, and $O$, respectively. For the input activations $I \in R^{N \times C_1 \times H \times W}$, $N$, $C_1$, $H$, and $W$ refer to the batch size, the number of input channels, the height, and width of the input feature maps, respectively.  For the weights $K \in R^{C_2 \times C_1 \times h \times w}$, $C_2$, $h$ and $w$ refer to the number of output channels, the height and width of the weights, respectively. For the output activations $O \in R^{N \times C_2 \times H \times W}$, it is computed as the convolution between the input activations $I$ and the weights $K$. In particular, every individual value of the output activations ${O_{[l,p,m,n]}}$ is calculated as follows.
\begin{equation}
\begin{aligned}
&{O_{[l,p,m,n]}} = \text{Convolution}(I,K) \\
&= \sum\limits_{o = 1}^{{C_1}} {\sum\limits_{j = 1}^{h - 1} {\sum\limits_{k = 1}^{w - 1} {{I_{[l,o,m'+j,n'+k]}} \times {K_{[p,o,j,k]}}} } } 
\end{aligned}
\end{equation}
where $l=0,...,N-1$, $m=0,...,H-1$, $n=0,...,W-1$, $o=0,...,C_1-1$, $p=0,...,C_2-1$, $m'=m-\frac{h-1}{2}$, $n'=n-\frac{w-1}{2}$. 

To apply the attention mechanism on the input activations $I$, previous attentional activations-based models produce the channel attention maps $A_c\in R^{N \times C_1 \times 1 \times 1}$ and spatial attention maps $A_s\in R^{N \times 1 \times H \times W}$ separately. For example, applying the channel attention maps $A_c$ on the input activations $I$ is presented as $O = \text{Convolution}((I \odot {A_c}),K)$, where $\odot$ refers to the Hadamard product and broadcasting during element-wise multiplication is omitted.

\subsubsection{Approximation problem of the attention maps} Instead of directly computing the three-dimensional attention map (N is omitted, otherwise the attention maps are of four dimensions.), all the current attentional activations-based models produce the attention maps separately into the channel attention maps $A_c$ and spatial attention maps $A_s$, which leads to the approximation problem of attention maps. However, to thoroughly attend the input activations $I$, we need to compute the attention maps $A_f \in R^{N \times C_1 \times H \times W}$ and apply it as $O = \text{Convolution}((I \odot {A_f}),K)$, which requires too much computational and parameter overhead.

Inspired by convolution, we adopt local connection and attention maps sharing to reduce the size of the attention maps. We compute the attention maps $A_a \in R^{N \times C_1 \times h \times w}$ as follows, where $\otimes$ is a special element-wise multiplication since it only works associated with convolution.
\begin{equation}
\begin{aligned}
&{O_{[l,p,m,n]}} = \text{Convolution}(I \otimes A_a,K) \\
&= \sum\limits_{o = 1}^{{C_1}} {\sum\limits_{j = 1}^{h - 1} {\sum\limits_{k = 1}^{w - 1} {( {{I_{[l,o,m'+j,n'+k]}}} \times {{A_a}_{[l,o,j,k]}} ) \times {K_{[p,o,j,k]}}} } }
\end{aligned}
\label{eq2}
\end{equation}
\paragraph{Insufficient capacity problem of the attention maps} To compute different channels of the output activations of the convolution, the input activations are constrained to be recalibrated by the same attention maps, which indicates the insufficient capacity of the attention maps. As each channel of the feature maps is considered as a feature detector, different channels of the output activations of the convolution expect the input activations to be adapted by different attention maps.

Take two channels of output activations of a convolutional layer as an example, the two channels are responsible for recognizing rectangle shape and triangle shape, respectively. Thus, it is reasonable for the two channels to expect that there are different attention maps for attending the input activations of the convolution (i.e., the attention maps to compute the channel of recognizing the rectangle shape should be different from the attention maps to compute the channel of recognizing the triangle shape). To meet this expectation, we need to compute the attention maps $A_{ic}\in R^{N \times C_2 \times C_1 \times 1 \times 1}$ and apply it on the input activations as follows.
\begin{equation}
\begin{aligned}
&{O_{[l,p,m,n]}} = \text{Convolution}(I \odot {{A_{ic}}_{[l,p,:,:,:]}},K) \\
&= \sum\limits_{o = 1}^{{C_1}} {\sum\limits_{j = 1}^{h - 1} {\sum\limits_{k = 1}^{w - 1} {( {{I_{[l,o,m'+j,n'+k]}}} \times {{A_{ic}}_{[l,p,o,0,0]}} ) \times {K_{[p,o,j,k]}}} } }
\end{aligned}
\label{eq3}
\end{equation}
To solve the approximation problem and the insufficient capacity problem of the attention maps together (i.e., combining the solution of Equation.\ref{eq2} and the solution of Equation.~\ref{eq3}), we introduce our proposed attention module by developing the AW-convolution. Specifically, we propose to compute the attention maps $A \in R^{N \times C_2 \times C_1 \times h \times w}$ and apply it as follows where the attention maps ${A_{[l,:,:,:,:]}}$ has the same shape as that of the weights instead of the input activations. In this paper, "Attentional weights" refers to the element-wise multiplication result between the attention maps and the weights. Similarly, "Attentional activations" refers to the element-wise multiplication result between the attention maps and the activations in previous attentional activations-based models. Thus, $I \otimes A$ and ${A_{[l,:,:,:,:]}} \odot K$ represent the attentional activations and attentional weights, respectively. To reduces half the number of element-wise multiplications, we calculate attentional weights instead of attentional activations as follows.
\begin{equation}
\begin{aligned}
&{O_{[l,p,m,n]}} = \text{Convolution}(I \otimes A,K)\\
&= \sum\limits_{o = 1}^{{C_1}} {\sum\limits_{j = 1}^{h - 1} {\sum\limits_{k = 1}^{w - 1} { {{I_{[l,o,m'+j,n'+k]}}} \times ( {A_{[l,p,o,j,k]}}  \times {K_{[p,o,j,k]}}} ) } } \\
&= \text{Convolution}(I, {A_{[l,:,:,:,:]}} \odot K) \\
&=\text{AW-Convolution}(I,A \odot K)
\end{aligned}
\end{equation}
\begin{figure*}[h]
\centering
\begin{subfigure}[b]{0.78\textwidth}
\centering
\includegraphics[width=1.0\linewidth]{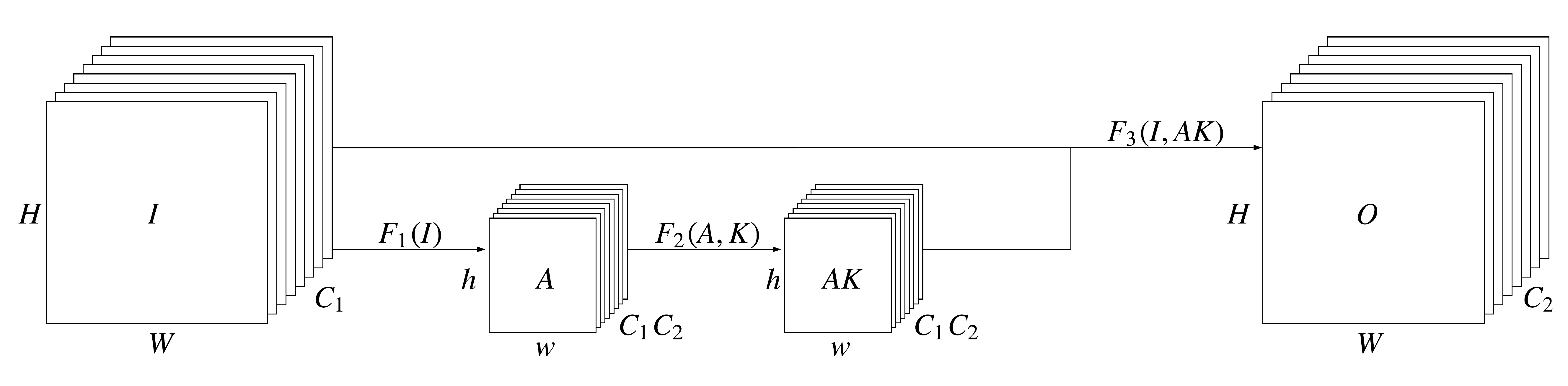}
   \caption{The AW-convolution architecture.}
\label{aw architecture}
\end{subfigure}
\begin{subfigure}[b]{0.78\textwidth}
\centering
\includegraphics[width=1.0\linewidth]{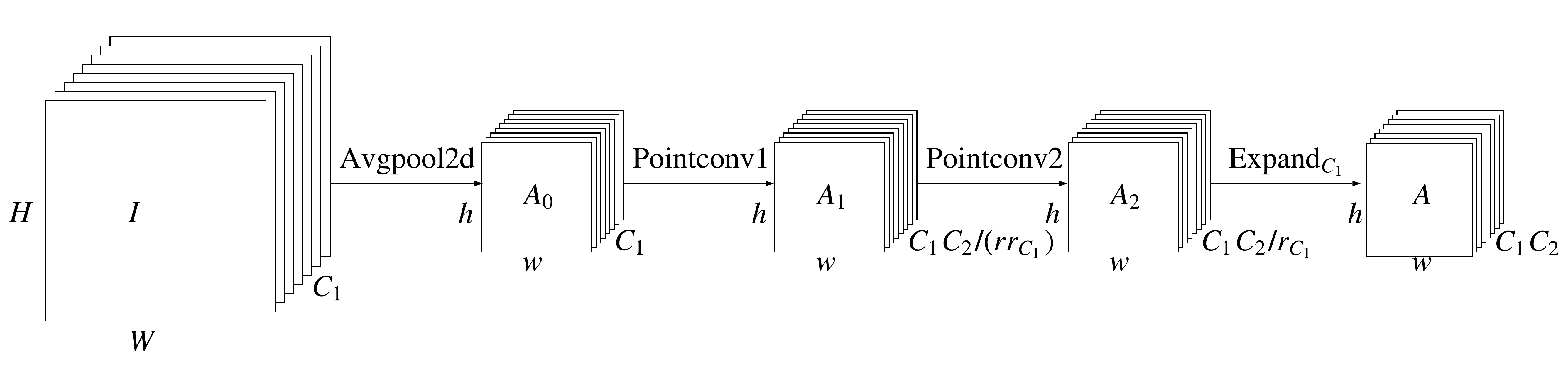}
   \caption{The architecture of calculating attention maps $A$.}
\label{F1 architecture}
\end{subfigure}
\caption{The architecture of our proposed attention module.}
\label{fig3}
\end{figure*}

\subsection{AW-convolution in proposed attention module}
The AW-convolution in our proposed attention module is presented in Figure~\ref{aw architecture}. In this figure, the attention maps $A$ has five dimensions, which is computed from the input activations $I$ as $A = F_1(I)$. $F_1$ is a function to calculate the attention maps $A$ given the input activations $I$. Then, the attentional weights $AK\in R^{N \times C_2 \times C_1 \times h \times w}$ is calculated as $AK = F_2(A, K) = K + A \odot K$. $F_2$ is a function to calculate the attentional weights $AK$ given the weights $K$ and the attention maps $A$. Finally, the output activations $O$ is calculated from the input activations $I$ and the attentional weights $AK$ as follows. 
\begin{equation}
\begin{aligned}
&{O_{[l,p,m,n]}} = {F_3}(I,AK) \\
&=\text{AW-Convolution}(I,AK)\\
& = \sum\limits_{i = 1}^{{C_1}} {\sum\limits_{j = 1}^{h - 1} {\sum\limits_{k = 1}^{w - 1} {{I_{[l,o,m'+j,n'+k]}} \times A{K_{[l,p,o,j,k]}}} } } \\
&= \text{Convolution}(I,A{K_{[l,:,:,:,:]}})
\end{aligned}
\end{equation}
where $F_3$ is a function to calculate the output activations $O$ given the input activations $I$ and the attentional weights $AK$. Compared with the traditional convolution, the attentional weights $AK$ of the AW-convolution in our proposed attention module has five dimensions rather than four dimensions, which are different from each other for every individual sample of the input activations batch to convolute.

It is also worth explaining the definition of the function $F_2$. $AK = K + A \odot K$ instead of  $AK = A \odot K$ is used to describe the function $F_2$ since it can be regarded as a residual design as follows.
\begin{equation}
\begin{aligned}
&O = {F_3}(I,AK) \\
&= \text{AW-Convolution}(I,{F_2}(A,K)) \\
&= \text{Convolution}(I,K) + \text{AW-Convolution}(I,A \odot K)
\end{aligned}
\end{equation}
\subsection{Calculating the attention maps $A$}
As shown in Figure~\ref{F1 architecture}, the architecture to compute the attention maps $A$ (i.e., the definition of the function $F_1$) is presented, which can be expressed as follows. Avgpool2d aggregates feature responses from the whole spatial extent and embeds them into $A_0$, and Pointconv1 and Pointconv2 followed by Relu redistribute the pooled information to capture the dynamic and no-linear dependencies between channels and spatial spaces. 
\begin{equation}
\begin{aligned}
&A = {F_1}(I) = \text{Expand}{_{{C_1}}}({A_2}) \\
&= \text{Expand}{_{{C_1}}}(\text{Pointconv2}({A_1})) \\
&= \text{Expand}{_{{C_1}}}(\text{Pointconv2}(\text{Pointconv1}({A_0}))) \\
&= \text{Expand}{_{{C_1}}}(\text{Pointconv2}(\text{Pointconv1}(\text{Avgpool2d}(I))))
\end{aligned}
\end{equation}
where Pointconv1 and Pointconv2 are pointwise convolutions. We add Batch Normalization and Relu layers after Pointconv1, while adding Batch Normalization and Sigmoid layers after Pointconv2, and they are omitted here to provide a clear expression. 

In Figure~\ref{F1 architecture}, Expand function along $C_1$ dimension, denoted as $\text{Expand}_{C_1}$, is used as an example, and Expand function can be also executed along $N$, $C_2$, $h$, and $w$ dimensions in a similar way. $\text{Expand}{_{C_1}}$ function is used to expand the tensor $A_2\in R^{N \times (C_2C_1/r_{C_{1}}) \times h \times w}$ into the attention maps $A\in R^{N \times C_2 \times C_1 \times h \times w}$ with the reduction ratio $r_{C_1}$, including necessary squeeze, reshape, and expand operations. $\text{Expand}_{C_1}$ can be expressed as follows.
\begin{equation}
\begin{aligned}
&A = \text{Expand}{_{{C_1}}}({A_2})\\
&= {A_2}.\text{reshape}(N,{C_2},{C_1}/{r_{C_1}},h,w)\\
&.\text{unsqueeze(dim=3)}\text{}\\
&\text{.expand}(N,{C_2},{C_1}/{r_{C_1}},{r_{C_1}},h,w)\text{}\\
&\text{.reshape}(N,{C_2},{C_1},h,w)
\end{aligned}
\end{equation}
Calculating the five-dimension attention maps $A$ is not an easy computational task without careful design. Thus, we analyze the additional computational complexity of an AW-convolution compared with a traditional convolution as a reference to refine this design. Considering the trade-off between computational complexity and accuracy, all the experiments in the remainder of this paper use the same settings for the architecture of calculating the attention maps $A$ in our proposed attention module, including $r_{C1} = {C_1}$, $r_{C_2}$ = $r_{hw}$ = 1, $r$ = 16, used in all the stages, and $AK = K + A \odot K$ as the definition for the function $F_2$. The details of refining the architecture of calculating the attention maps $A$ are in Section C of Supplementary Material.

\subsection{Integrating with other attention-based modules}
\begin{figure}[h]
\centering
\begin{subfigure}{0.40\textwidth}
\centering
\includegraphics[width=1.0\linewidth]{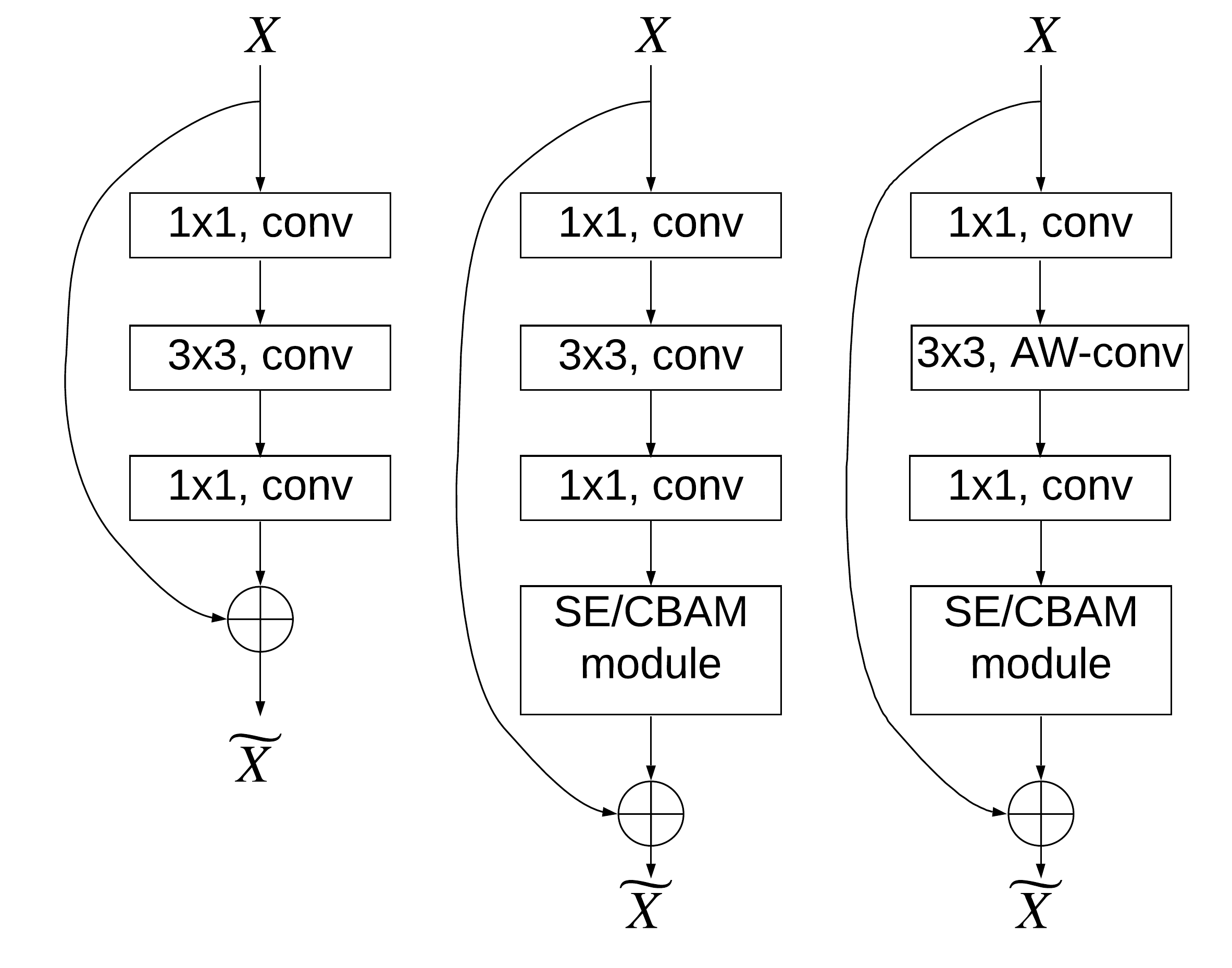}
\caption{Integrating with SE-ResNet/CBAM-ResNet.}
\label{integrating}
\end{subfigure}
\begin{subfigure}{0.40\textwidth}
\centering
\includegraphics[width=1.0\linewidth]{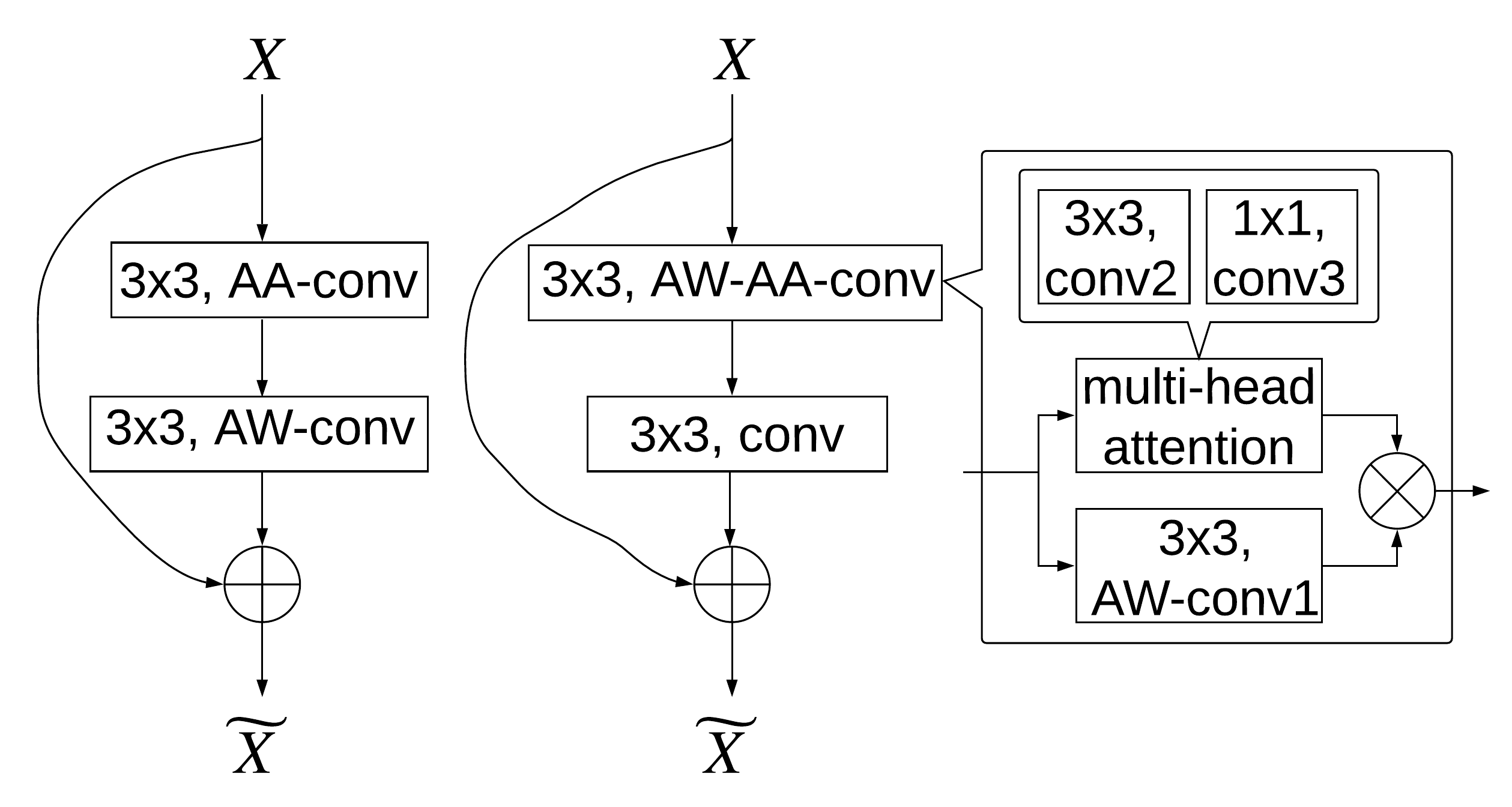}
\caption{Integrating with AA-Wide-ResNet.}
\label{integratingaa}
\end{subfigure}
\caption{The schema of bottlenecks and blocks when integrating with our proposed attention module.}
\label{fig4}
\end{figure}

In this section, we show how to integrate our proposed attention module with the previous attention-based convolutional neural networks to demonstrate the complementary relationship between our proposed attention module and other attention-based modules. Since applying our proposed attention module is using the AW-convolution to replace the traditional convolution, we can easily integrate our proposed attention module with any convolutional neural networks consisting of traditional convolution, including all the recently developed attention-based models \cite{hu2018squeeze, woo2018cbam,park2018bam,linsley2018learning,Bello_2019_ICCV}.

We choose the recent attentional activations-based models, i.e., SE-Net and CBAM-Net, as examples to show how to integrate our proposed attention module with other attention-based models. Here we use the popular ResNet \cite{he2016deep} as the backbone to apply the attention mechanism. As shown in Figure~\ref{integrating}, the left side is the structure of a primary bottleneck in ResNet. The middle one is the structure of a bottleneck with SE/CBAM modules in SE-ResNet/CBAM-ResNet. Integrating the central bottleneck with our proposed attention module is completed by replacing its $3 \times 3$ convolution with a $3 \times 3$ AW-convolution, and its final structure in AW-SE-ResNet/AW-CBAM-ResNet is shown on the right side. In summary, our proposed attention module is a general module to be integrated seamlessly with any CNNs architectures, including previous attention-based CNNs.

In Figure~\ref{integratingaa},  we integrate our proposed attention module with Attention Augmented (AA) convolutional networks and describe the architecture of their possible AW-AA-blocks. Experiments of AW-AA-Wide-ResNet \cite{BMVC2016_87,Bello_2019_ICCV} on CIFAR-100 image classification \cite{Krizhevsky09}, as shown in Table~\ref{table3}, suggest that integrating our proposed attention module with attention-based models should be explored carefully since different integration architectures achieve different accuracy. In some cases, an improper integration architecture leads to a small accuracy drop. With a careful design for integration, our proposed attention module is complementary to AA convolution and improves the accuracy of AA-Net further. 

In particular, we augment Wide-ResNet-28-10 by replacing the first convolution of all the residual blocks with Attention Augmented convolution, which is the AA-Wide-ResNet baseline. Here we set $N_h=8$ heads, $k=2v=0.2$, and a minimum of $20$ dimensions per head for the keys as in~\cite{Bello_2019_ICCV}. In this figure, we can develop four possible architectures to integrate our proposed attention module with AA-Wide-ResNet. On the left side of this figure, we can build AA-Wide-ResNet-0 by replacing the second $3 \times 3$ convolution with our AW-convolution in an AA-block. On the right side, there are three traditional convolutions in an AA convolution, including conv1, conv2, conv3. The conv1 is parallel to the multi-head attention, while the conv2 and conv3 are used to calculate QKV (i.e., queries, keys, and values) and to output attention maps, respectively. On the medial side, we can replace one traditional convolution from conv1, conv2, or conv3 with our AW-conv1, AW-conv2, or AW-conv3 to construct AW-AA-Wide-ResNet-1, AW-AA-Wide-ResNet-2, or AW-AA-Wide-ResNet-3, respectively. The Top-1 accuracy of AW-AA-Wide-ResNet-0 is $1.87\%$ higher than the AA-Wide-ResNet baseline, while AW-AA-Wide-ResNet-3 shows worse performance by $0.10\%$ drop.

\begin{table*}[h]
\begin{center}
\begin{tabular}{lllll}
\hline
Model & Top-1 Error & Top-5 Error  & GFLOPs  & Parameters (M)   \\
\hline
ResNet50 Baseline \cite{he2016deep} & $24.56\%$(+$0.00\%$) & $7.50\%$ & $3.86$ &  $25.56$  \\
AW-ResNet50 & $23.38\%$(+$1.18\%$) & $6.79\%$ & $3.87$ &  $25.72$\\
SE-ResNet50 \cite{hu2018squeeze} & $23.14\%$(+$1.42\%$) & $6.70\%$ & $3.87$ &  $28.09$\\
AW-SE-ResNet50 & $22.72\%$(+$1.84\%$) & $6.47\%$ & $3.88$ &  $28.25$\\
AW-CBAM-ResNet50 (MaxPool) & $22.82\%$(+$1.74\%$) & $6.41\%$ & $3.89$ &  $28.25$\\
AW-CBAM-ResNet50 (Spatial) & $23.20\%$(+$1.36\%$) & $6.58\%$ & $3.90$ &  $28.25$\\
\hline
ResNet101 Baseline \cite{he2016deep} & $23.38\%$(+$0.00\%$) & $6.88\%$ & $7.57$ &  $44.55$  \\
AW-ResNet101 & $22.38\%$(+$1.00\%$) & $6.21\%$ & $7.58$ &  $44.95$\\
SE-ResNet101 \cite{hu2018squeeze} & $22.35\%$(+$1.03\%$) & $6.19\%$ & $7.58$ &  $49.33$\\
AW-SE-ResNet101 & $21.78\%$(+$1.60\%$) & $5.74\%$ & $7.59$ &  $49.73$\\
AW-CBAM-ResNet101 (MaxPool) & $21.64\%$(+$1.74\%$) & $5.76\%$ & $7.60$ &  $49.73$\\
AW-CBAM-ResNet101 (Spatial) & $22.32\%$(+$1.06\%$) & $6.18\%$ & $7.61$ &  $49.73$\\
\hline
MobileNet Baseline \cite{DBLP:journals/corr/HowardZCKWWAA17} & $31.39\%$(+$0.00\%$) & $11.51\%$ & $0.569$ &  $4.23$  \\
SE-MobileNet \cite{hu2018squeeze} & $29.97\%$(+$1.42\%$) & $10.63\%$ & $0.581$ &  $5.07$\\
AW-SE-MobileNet & $29.41\%$(+$1.98\%$) & $10.59\%$ & $0.623$ &  $5.52$\\
CBAM-MobileNet \cite{woo2018cbam} & $29.01\%$(+$2.38\%$) & $9.99\%$ & $0.611$ &  $5.07$\\
AW-CBAM-MobileNet (Spatial) & $28.82\%$(+$2.57\%$) & $9.98\%$ & $0.652$ &  $5.52$\\
\hline
\end{tabular}
\end{center}
\caption{Comparisons of attention-based models on ImageNet classfication.}
\label{table5}
\end{table*}

\section{Experimental results}
In this section, we use extensive experiments to demonstrate the effectiveness of our proposed attention module. We use ResNet \cite{he2016deep}, MobileNet \cite{DBLP:journals/corr/HowardZCKWWAA17}, SSD300 \cite{liu2016ssd}, and Faster R-CNN \cite{ren2015faster} as the baseline models, and various variants of these models are developed, including using our proposed attention module for these baseline models and integrating our proposed attention module with their attentional activations-based models. The datasets to train these models include CIFAR-100 classification \cite{Krizhevsky09}, ImageNet classification \cite{deng2009imagenet}, VOC object detection datasets \cite{everingham2015pascal} (The experimental results are included in Section E of Supplementary Material.), and COCO object detection datasets \cite{lin2014microsoft}. To have a better interpretation of our proposed attention module, its feature visualizations using Grad-CAM \cite{selvaraju2017grad} can be found in Section F of Supplementary Material. All the data augmentation and training settings can be found in Section A of Supplementary Material.

\subsection{ImageNet image classification}
To investigate the performance of our proposed attention module on high-resolution images, we train ResNet50 and ResNet101 \cite{he2016deep} and their attention-based variants on the ImageNet classification dataset \cite{deng2009imagenet}. According to the results shown in Table~\ref{table5}, our proposed attention module is complementary to other attentional activations-based models. AW-ResNet50 achieves a $1.18\%$ Top-1 error reduction compared with the ResNet50 baseline. Integrating with our proposed attention module, SE-ResNet50 \cite{hu2018squeeze} can improve further by $0.42\%$ Top-1 accuracy. The Top-1 accuracy of our AW-SE-ResNet101 is $1.60\%$ and $0.57\%$ higher than that of ResNet101 and SE-ResNet101, respectively. To integrate with CBAM-ResNet \cite{woo2018cbam} more carefully, we define CBAM-ResNet (MaxPool) and CBAM-ResNet (Spatial) separately. In CBAM-ResNet (MaxPool), we do not deploy the spatial attention maps, while we do not use max-pooled features in CBAM-ResNet (Spatial). The Top-1 accuracy of AW-CBAM-ResNet50 (MaxPool) and AW-CBAM-ResNet50 (Spatial) are better than AW-ResNet50 by $0.56\%$ and $0.18\%$, respectively, but worse than AW-SE-ResNet50. The number of additional parameters for our proposed attention module is $0.16$ M, which is much smaller than $2.83$ M (i.e., one-sixteenth) of SE and CBAM modules. Moreover, it takes only $0.01$ GFLOPs to apply our proposed attention module on the ResNet50 model on ImageNet classification, which is comparable with $0.01$ GFLOPs and $0.04$ to adopt the SE and CBAM modules and is negligible in terms of FLOPs to implement the baseline model. The computational complexity analysis introduced by the attention mechanism can be found in Section B of Supplementary Material.

\begin{table*}[h]
\begin{center}
\begin{tabular}{lllll}
\hline
Model & Top-1 Error & Top-5 Error  & GFLOPs  & Parameters (M)   \\
\hline
AA-Wide-ResNet Baseline \cite{Bello_2019_ICCV} & $28.01\%$($+0.00\%$) & $7.92\%$ & $3.89$ &  $8.43$  \\
AW-AA-Wide-ResNet-0  & $26.14\%$($+1.87\%$) & $7.43\%$ & $3.90$ &  $8.50$\\
AW-AA-Wide-ResNet-1  & $27.17\%$($+0.84\%$) & $7.49\%$ & $3.89$ &  $8.48$\\
AW-AA-Wide-ResNet-2 & $27.09\%$($+0.92\%$) & $8.00\%$ & $3.89$ &  $8.45$\\
AW-AA-Wide-ResNet-3  & $28.11\%$($-0.10\%$) & $8.24\%$ & $3.89$ &  $8.44$\\
\hline
ResNet50 Baseline \cite{he2016deep} & $22.33\%$($+0.00\%$) & $5.83\%$ & $1.22$ &  $23.71$  \\
AW-ResNet50 & $19.87\%$($+2.46\%$) & $4.76\%$ & $1.23$ &  $23.87$\\
SE-ResNet50 \cite{hu2018squeeze} & $20.43\%$($+1.90\%$) & $5.01\%$ & $1.23$ &  $26.24$\\
AW-SE-ResNet50 & $19.00\%$($+3.33\%$) & $4.51\%$ & $1.24$ &  $26.40$\\
CBAM-ResNet50 \cite{woo2018cbam} & $19.46\%$($+2.87\%$) & $4.56\%$ & $1.24$ &  $26.24$\\
AW-CBAM-ResNet50 & $18.94\%$($+3.39\%$) & $4.76\%$ & $1.25$ &  $26.40$\\
\hline
\end{tabular}
\end{center}
\caption{Comparisons of attention-based models on CIFAR-100 classfication.}
\label{table3}
\end{table*}
\begin{table*}[h]
\begin{center}
\begin{tabular}{lllll}
\hline
Backbone & Detector  &  mAP@[0.5, 0.95] & mAP@0.5 &  mAP@0.75      \\
\hline
ResNet101-FPN \cite{lin2017feature} & Faster R-CNN  & $37.13$($+0.00\%$) & $58.28$ &  $40.29$  \\
ResNet101-AW-FPN & Faster R-CNN  & $37.76$($+0.63\%$) & $59.17$ &  $40.91$\\
ResNet101-SE-FPN \cite{hu2018squeeze}  & Faster R-CNN  & $38.11$($+0.98\%$) & $59.41$ &  $41.33$\\
ResNet101-AW-SE-FPN & Faster R-CNN  & $38.45$($+1.32\%$) & $59.70$ &  $41.86$\\
ResNet101-CBAM-FPN \cite{woo2018cbam} & Faster R-CNN   & $ 37.74$($+0.61\%$) & $58.84$ &  $40.77$\\
ResNet101-AW-CBAM-FPN & Faster R-CNN  & $38.19$($+1.06\%$) & $59.52$ &  $41.43$\\
\hline
\end{tabular}
\end{center}
\caption{Comparisons of attention-based Faster R-CNN on COCO.}
\label{table8}
\end{table*}
\subsubsection{Resource-constrained architecture} 
Driven by demand for mobile applications, many depthwise convolution-based models are developed to take care of the trade-off between accuracy and efficiency. To inspect the generalization of our proposed attention module in this resource-constrained scenario, we conduct the ImageNet classification \cite{deng2009imagenet} with the MobileNet architecture \cite{DBLP:journals/corr/HowardZCKWWAA17}. Since our proposed attention module is efficient in terms of both the storage and computational complexity, integrating it into the light-weight architecture is worth exploring. We apply our proposed attention module to pointwise convolution instead of depthwise convolution in every two depthwise separable convolutions. When integrating with the CBAM models \cite{woo2018cbam}, we remove the max-pooled features and keep spatial attention maps. As shown in Table~\ref{table5}, AW-SE-MobileNet and AW-CBAM-MobileNet achieve $0.56\%$ and $0.19\%$ Top-1 accuracy improvements compared with SE-MobileNet \cite{hu2018squeeze} and CBAM-MobileNet, respectively. It is an impressive result that the Top-1 accuracy of AW-CBAM-MobileNet is $2.57\%$ better than that of the MobileNet baseline. For the MobileNet model, our proposed attention module increases the computation by $0.041$ GFLOPs, while SE and CBAM modules need $0.012$ and $0.041$ GFLOPs, respectively. Also, the required parameters for our proposed attention module are $0.45$ M, which is much less than $0.84$ M for SE and CBAM modules.

\subsection{CIFAR-100 image classification}
According to the results shown in Table~\ref{table3}, we conclude that our proposed attention module can boost the CIFAR-100 \cite{Krizhevsky09} accuracy of both the ResNet50 baseline model \cite{he2016deep} and their attentional activations-based models with negligible additional computational complexity. AW-ResNet50 achieves a $2.46\%$ Top-1 error reduction compared with the ResNet50 baseline. Integrating with our proposed attention module, SE-ResNet50 \cite{hu2018squeeze} and CBAM-ResNet50 \cite{woo2018cbam} can increase Top-1 accuracy by $1.43\%$ and $1.52\%$, respectively. In terms of the computational complexity, our proposed attention module requires $0.01$ GFLOPs, which is acceptable compared with $0.01$ GFLOPs for the SE module, $0.02$ GFLOPs for the CBAM module, and $1.22$ GFLOPs for the baseline model. Besides, we only introduce $0.16$ M parameters for our proposed attention module, which is less than $2.53$ M parameters for the SE and CBAM modules. We train ResNet with various depths on CIRAR-100 image classification to show that our proposed attention module works for CNNs with different depths, which are in Section D of Supplementary Material.

\subsection{Object Detection on COCO}
To show the generalization of our proposed attention module, we apply it to object detection tasks. We evaluate our proposed attention module further on the COCO dataset \cite{lin2014microsoft}, which contains $118K$ images (i.e., train2017) for training and $5K$ images (i.e., val2017) for validating. We use Faster R-CNN \cite{ren2015faster} as the detection method with the ResNet101-FPN backbone \cite{lin2017feature}. Here we intend to evaluate the benefits of applying our proposed attention module on the ResNet101-FPN backbone \cite{lin2017feature}, where all the lateral and output convolutions of the FPN adopt our AW-convolution. The SE and CBAM modules are placed right before the lateral and output convolutions. As shown in Table~\ref{table8}, applying our proposed attention module on ResNet101-FPN boosts mAP@[0.5, 0.95] by $0.63$ for the Faster R-CNN baseline. Integrating with attentional activations-based models, Faster R-CNNs with the backbones of ResNet101-AW-SE-FPN and ResNet101-AW-CBAM-FPN outperform Faster R-CNNs with the backbones of  ResNet101-SE-FPN and ResNet101-CBAM-FPN by $0.34$ and $0.45$ on COCO’s standard metric AP. 

\section{Conclusion}
In this paper, We analyze the two ignored problems in attentional activations-based models: the approximation problem and the insufficient capacity problem of the attention maps. To address the two problems together, we propose an attention module by developing the AW-convolution, where the shape of the attention maps matches that of the weights rather than the activations, and integrate it with attention-based models as a complementary method to enlarge their attentional capability. We have implemented extensive experiments to demonstrate the effectiveness of our proposed attention module, both on image classification and object detection tasks. Our proposed attention module alone shows noticeable accuracy improvement compared with baseline models. More importantly, integrating our proposed module with previous attention-based models, such as AA-Net \cite{Bello_2019_ICCV}, SE-Net \cite{hu2018squeeze}, and CBAM-Net \cite{woo2018cbam}, will further boost their performance. 

\clearpage
\bibliography{aaai}

\end{document}